\pdfoutput=1

\documentclass[11pt]{article}

\usepackage[]{acl}

\usepackage{times}
\usepackage{latexsym}
\usepackage{amsmath}
\usepackage{graphicx}
\usepackage{makecell}
\usepackage{arydshln}
\usepackage{multirow}
\usepackage{caption}
\usepackage{subcaption}
\usepackage{amssymb}
\usepackage[T1]{fontenc}
\usepackage[utf8]{inputenc}
\usepackage{microtype}

\title{Adapting to Non-Centered Languages for Zero-shot Multilingual Translation}

\author{Zhi Qu \and Taro Watanabe \\
        Nara Institute of Science and Technology\\
        \{qu.zhi.pv5, taro\}@is.naist.jp}

\begin{document}
\maketitle
\begin{abstract}
Multilingual neural machine translation can translate unseen language pairs during training, i.e. zero-shot translation.
However, the zero-shot translation is always unstable.
Although prior works attributed the instability to the domination of central language, e.g. English, we supplement this viewpoint with the strict dependence of non-centered languages.
In this work, we propose a simple, lightweight yet effective language-specific modeling method by adapting to non-centered languages and combining the shared information and the language-specific information to counteract the instability of zero-shot translation.
Experiments with Transformer on IWSLT17, Europarl, TED talks, and OPUS-100 datasets show that our method not only performs better than strong baselines in centered data conditions but also can easily fit non-centered data conditions.
By further investigating the layer attribution, we show that our proposed method can disentangle the coupled representation in the correct direction.\footnote{Codes and detailed results are available in:  \url{https://github.com/zhiqu22/AdapNonCenter}}

\end{abstract}

\section{Introduction}
Training multilingual neural machine translation (MNMT) system requires enormous number of parameters and resources, but the zero-shot translation, namely translating unseen language pairs during training, has shown the potential to simplify the MNMT \cite{firat-2017}.
\citet{Johnson} has shown that adding language tokens, e.g. <en>, at the beginning of a sentence allows the model to build cross-linguistic representation by treating the token as translation instruction specifying target language.
However, the zero-shot translation is always unstable.
One possibility causing the instability of zero-shot translation is spurious correlation \cite{gu-2019}.
The target linguistic representation captured by the model is directly and strictly dependent on encoded source linguistic information instead of learning specific representations for source and target language, then combining independent linguistic representations to generate results.
Prior works \cite{lakew-2019,fan,rios-2020,freitag-2020,liu-2021} indicated that the spurious correlation is caused by the centered data condition in which multilingual data is constructed by bridging a central language, e.g. English, to other non-centered languages.
The central language will dominate the representation in the MNMT model to degenerate the information specific to non-centered languages since multilingual data comprises a set of bilingual data constructed by coupling non-centered languages with the central language.
However, the non-centered data condition without any central language is also unstable in zero-shot translation.\footnote{We give specific examples in Section \ref{section:data}.}
Therefore, simply attributing the instability of zero-shot translation to the central language cannot fit all cases of zero-shot translation.

We move the perspective from the domination of the central language to the weakness of non-centered languages.
The problems of zero-shot translation could be attributed to the strict dependence of non-centered languages.
Specifically, a non-centered language would strictly depend on another language as a strongly related language pair to prohibit learning robust and independent translation instructions for zero-shot translation.
Under this hypothesis, the centered data condition is a special case of this description, because all non-centered languages depend on the central language.
In this light, a key to improving zero-shot translation is disentangling non-centered languages from the strict dependence which is built in training.

Specifically, we model extra language-specific (LS) components \cite{sachan-2018,philip-2020,escolano-2021,zhang-2021} adapting to non-centered languages in a mixing shared and LS information mode \cite{zhang-2021}, our objective is to enhance the weak representations for assisting the balance of cross-linguistic representation in shared information container to improve the quality of translation \citet{cheng-2022, shao-2022}.
Furthermore, the mixing mode can decrease the complexity of LS modeling since we treat the representation space of the MNMT model as the combination of shared and LS information, and we no longer build the independent representation space for each language \cite{sachan-2018, escolano-2021}.
In this motivation, we propose a simple, lightweight yet effective method to augment feed-forward network of Transformer \cite{attention} by LS components adapting to non-centered languages.

Our contributions are as follows:

\begin{itemize}
  \item Our lightweight method achieves considerable gains on multilingual and zero-shot translation and performs stably in IWSLT17, Europarl, TED talks and OPUS-100.
  \item We describe the strict dependence of non-centered languages to supplement the prior viewpoint of zero-shot translation, and verify it by experiments under different data conditions with and without the central language.
  \item Our work explores decreasing complexity in LS modeling.
  We also through the analysis via layer attribution \cite{attribution} to show the significance of our methods in decoupling representations of MNMT.
\end{itemize}

\section{Related Work}
Initially, \citet{Johnson} laid the foundation of zero-shot translation which endorses training the MNMT model under the centered data condition and put forward the thinking about the instability of data conditions on zero-shot translation.
On this basis, \citet{gu-2019} also showed that the performance of zero-shot translation is sensitive to parameters for initialization, which is another cause of instability.
In this paper, we systematically described this instability (Section \ref{section:instability}) and tested it experimentally.

In the early stages, \citet{Mattoni-2017} pointed out that increasing corpus size can effectively improve zero-shot translation.
However, the spurious correlation \cite{gu-2019} means that unreasonably increasing training data could degenerate the zero-shot translation due to strict dependence.
Since then, the concern of the centered data condition was started to be discussed by several different strategies.
\citet{fan, freitag-2020} augmented the training data to make all languages interconnected, which will result in an excessive increase in training costs.
\citet{lakew-2019} explored incrementally training the MNMT model by monolingual data, and 
\citet{gu-2019, zhang-2020} generated synthetic data for the zero-shot directions by backtranslation.
These methods transformed the zero-shot task to zero-resource task.

Another line of work on improving zero-shot translation is to adjust the learning of representations in the MNMT model.
\citet{lu-2018, pham-2019, zhang-2020, liu-2021} focused on restricting the representation of encoder outputs to be language-agnostic, but the restriction may reduce the performance of the model trained by large-scale datasets.
\citet{pan-2021} aligned representations from different languages via contrastive learning and the additional dictionary.
\citet{philip-2020, yang-2021, zhang-2021} explored to enhance the influence of LS features in the translation.
Our work continues in this direction, but with a special focus on only enhancing the decoding step and mixing shared and LS information.

Our work is based on LS modeling which is the heuristic variation of Mixture-of-Experts model \cite{MoE}, because it aims to build extra components as experts to directionally improve linguistic features.
\citet{sachan-2018} and \citet{escolano-2021} built LS encoder or decoder, but multi-encoder/decoder architecture has too many parameters.
\citet{wang-2018} divided neural cells into LS parts and \citet{lin-2021} divided LS subnets from the model, but these methods limited the learning capacity.
\citet{bapna-2019} and \citet{philip-2020} added LS adapters on the end of encoder and decoder and fine tuned for LS representations.
\citet{zaremoodi-2018, zhang-2021} explored the paradigm of constructing LS components to assist the shared information.
However, extra components always increase the cost of modeling significantly when languages existed too much.
The investigation about the importance of LS information specified to target language \cite{Lee-2017, blackwood-2018, pham-2019, wu-2021} enlightens us to limit the improving LS information in the decoding process to achieve lightweight LS modeling.

\section{Central Language Aware Multilingual Neural Machine Translation}
We employ Transformer \cite{attention} as the backbone to construct our architecture.
Consider a set of $m$ languages $\mathbb{L}=\{l_1,l_2,\ldots,l_m\}$, we assign the first language ${l}_1$ as the central language~${l}_c$.
The non-centered set is the subset of $\mathbb{L}$, that is $\mathbb{L'}=\{l_2,l_3,\ldots,l_m\}$.
We follow prior works \cite{zhang-2020,liu-2021,wu-2021} to assign English as the center of multilingual data.
Given the original input sequence of symbol representation to the encoder $\boldsymbol{x}=x_1,x_2,...,x_i$ and the output sequence generated by decoder $\boldsymbol{y}=y_1,y_2,...,y_j$, we follow the method of \citet{Johnson} to insert the language token at the beginning of $\boldsymbol{x}$ as translation instruction.
Therefore, the actual input sequence is $\boldsymbol{x}'=(l, \boldsymbol{x})$, and we model the translation of $\boldsymbol{x}'$ to $\boldsymbol{y}$ with Transformer.
We only build LS layers (LSLs) parallel with the Feed-Forward Network (FFN) layers in the decoder of Transformer, and keep the self-attention and cross-attention mechanism fixed.

The FFN of transformer consists of two fully connected neural networks with a ReLU activation function in between:
\begin{equation}\label{eq1}
    \operatorname{FFN}(\boldsymbol{h})=\max(0, \boldsymbol{h} \mathbf{W}_1 + \boldsymbol{b}_1) \mathbf{W}_2 + \boldsymbol{b}_2
\end{equation}
Where $\boldsymbol{h}$ is the input vector, $\mathbf{W}$ indicates parameter matrices for projections, and $\boldsymbol{b}$ indicates bias parameter matrices.
LSLs are a series of neural networks specified to $\mathbb{L'}$. Each LSL is similar to FFN in architecture but can be relatively light in inner size:
\begin{equation}\label{eq2}
    \operatorname{LSL}_l(\boldsymbol{h})=\max(0,\boldsymbol{h} \mathbf{W}_{1}^{l}+\boldsymbol{b}_{1}^{l}) \mathbf{W}_{2}^{l}+\boldsymbol{b}_{2}^{l}
\end{equation}
where $l \in \mathbb{L}'$.
The trade-off between shared and LS information is difficult \cite{zhang-2021,wang-2021}, because the information that each language carries is not absolutely equal.
To balance the shared and LS information, we introduce a set of learnable scalars in each decoder layer $\mathbb{T}=\{t_{l_2},t_{l_3},\ldots,t_{l_m}\}$.
Elements of $\mathbb{T}$ correspond to languages of $ \mathbb{L'} $ one by one, then each $t$ connects LS information with the shared information.
We initialize $t$ to 0.1, then parameters\footnote{we report the distribution of LS information weights for large-scale dataset (99 languages) in the Appendix \ref{appendix:dis}.} of $\mathbb{T}$ are updated during training together with other parameters.

\begin{figure}[!t]
    \centering
    \includegraphics[width=0.47\textwidth]{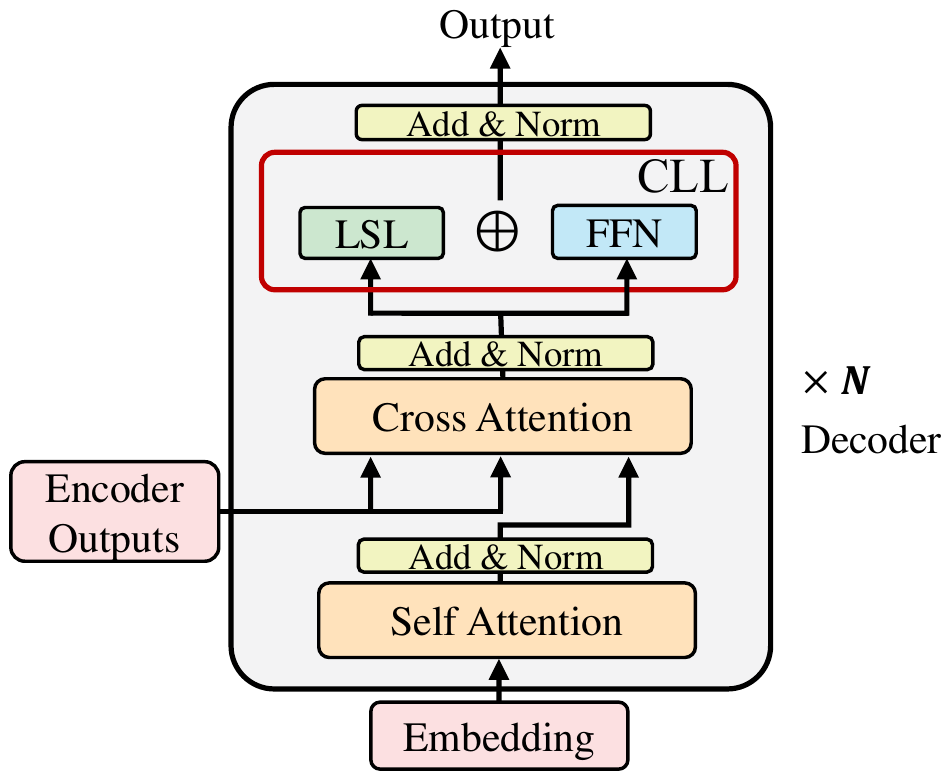}
    \caption{
    Illustrations of our proposed architecture modified from the decoder of Transformer. $\bigoplus$ indicates weighted plus, $N$ is the number of decoder layers.}
    \label{figure1}
\end{figure}

To differentiate the central language from non-centered languages, the original FFN of Transformer's decoder is used as the shared information space for all languages of $\mathbb{L}$, and we only construct lightweight LSLs to learn independent linguistic information for non-centered languages of $\mathbb{L'}$.
Therefore, the complete architecture of \textbf{C}entral \textbf{L}anguage-aware \textbf{L}ayer (CLL) is:

\begin{equation}\label{eq3}
    \operatorname{CLL}_l(\boldsymbol{h})= 
    \begin{cases}
        \operatorname{FFN}(\boldsymbol{h})+t_l\operatorname{LSL}_l(\boldsymbol{h})&l \in \mathbb{L'}\\\\
        \operatorname{FFN}(\boldsymbol{h})&l=l_c
    \end{cases}
\end{equation}
Based on the piecewise function Eq.(\ref{eq3}), the role of central language will be abandoned in non-centered data conditions, namely the case of $l=l_c$ will not be triggered.
We illustrate the architecture of CLL in Figure \ref{figure1}: The CLL is a component, including lightweight LSLs and FFN, to replace the original role of FFN in each decoder layer of Transformer.
Compared to the Mixture of Experts which is the generalization of the gating mechanism \cite{MoE}, a deterministic route specific to language replaces the gate in CLL.
For convenience, we use \textbf {FCLL} (full CLL) to indicate that the model in which all decoder layers are constructed in the form of our proposed architecture.

We introduce a variation named \textbf{SD} that constructs CLL in a single decoder layer among all layers of Transformer, namely Single-Disentangled CLL.
Inspired by the work of \citet{liu-2021}, we remove the residual connection of FFN in a middle encoder layer to weaken the linguistic features of encoding.
To keep the balance between weakening encoding and improving decoding, we empirically build CLL in the middle decoder.
Specifically, given $N$ encoder and decoder layers of Transformer, we remove the residual connection of the FFN in the encoder and replace the FFN with CLL in the decoder at $N/2+1^{th}$ layer of both networks.
Our experiments (Section \ref{section:result}) empirically show that SD has comparable performance with FCLL in small-scale datasets, although SD is more parameter-efficient than FCLL (Table~\ref{tab1}).

\begin{table}
\centering
\resizebox{0.47\textwidth}{!}{
\begin{tabular}{lll}
\hline
Method & +Params& Position\\
\hline
baseline & None  & None\\
FCLL & $\mathcal{O}(k)$  & Decoder\\
SD & $\mathcal{O}(k/N)$   & Decoder\\
\citet{philip-2020} & $\mathcal{O}(2k)$  & All\\
\citet{zhang-2021} & $\mathcal{O}(5k)$ & All\\
\citet{sachan-2018} & $\mathcal{O}(K)$ & Decoder\\
\hline
\end{tabular}}
\caption{Number of parameters required for different LS modeling methods. $N$, $k$ and $K$ denote the number of encoder/decoder layers, parameters per LS layer, and parameters per encoder/decoder layer ($k\ll K$), respectively. Position indicates the position of a model to construct LS components. }
\label{tab1}
\end{table}

\section{Experiments}\label{section:exp}

\subsection{Dataset}\label{section:data}
We take IWSLT17 \cite{IWSLT} and restrict 4 languages from MMCR4NLP \cite{mmcr4nlp} to verify basic abilities of multilingual and zero-shot translation.
We follow \citet{philip-2020} to experiment on TED talks \cite{qi-etal-2018} and restrict top 20 languages.
We also experiment on OPUS-100 \cite{zhang-2020} to exhaustively explore the capacity of our proposed method in the large-scale dataset.
English is the central language of those cases.

\begin{figure}[t]
    \centering
    \begin{subfigure}[b]{0.35\linewidth}
        \centering
        \includegraphics[width=\linewidth]{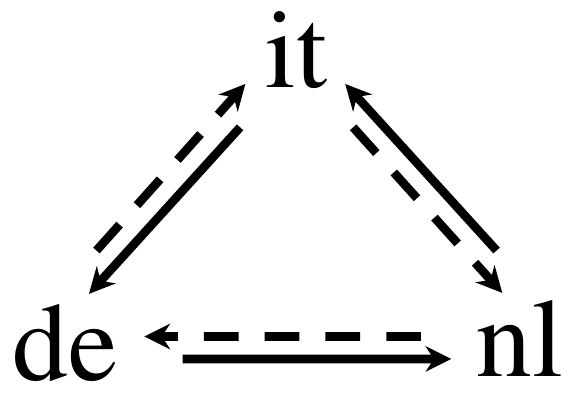}
        \caption{ }
        \label{figure2.1}
    \end{subfigure}
    \begin{subfigure}[b]{0.27\linewidth}
        \centering
        \includegraphics[width=\linewidth]{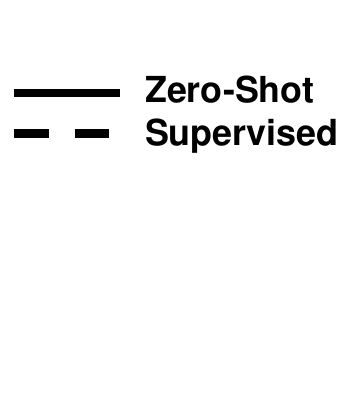}
    \end{subfigure}
    \begin{subfigure}[b]{0.35\linewidth}
        \centering
        \includegraphics[width=\linewidth]{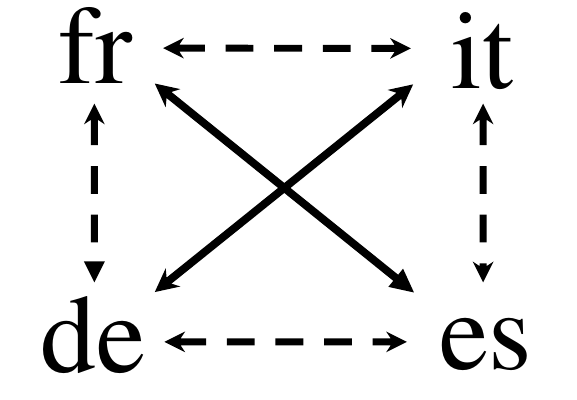}
        \caption{ }
        \label{figure2.2}
    \end{subfigure}
\caption{\label{figure2}
Illustration supervised and zero-shot directions for Triangle and Square cases.}
\end{figure}

To show the strict dependence of non-centered languages, we design two different cases without central language, namely all languages in the set are non-centered.
We extract and reorganize Europarl v7 \cite{europarl} from MMCR4NLP:
1) Triangle case, where each language appears at the target and source sides only once.
Our motivation is to build the strict dependence under the non-centered data condition, and each language pair has more training data than IWSLT.
Figure~\ref{figure2.1} shows its translation directions we designed.
2) Square case, that is designed for avoiding strict dependence as indicated in Figure \ref{figure2.2}.
Our motivation is to avoid completely interconnecting all languages \cite{fan, freitag-2020} while building balanced data conditions.

We list details of datasets in Appendix \ref{appendix:dataset}.
And all cases are evaluated via official test sets.

\subsection{Experimental Setup}
We employ Fairseq \cite{fairseq}, the open-source implementation, of Transformer \cite{attention} as backbone.
Generally, we apply the Moses tokenizer\footnote{\url{https://github.com/moses-smt/mosesdecoder}} for tokenization and detokenization, and use SentencePiece \cite{sentencepiece} to learn subword vocabulary.
Although the detail of training subword vocabulary for each case has differences, we always train the joint vocabulary including the source and target side, and set \emph{share-all-embedding} in Fairseq.
To prevent the unbalanced data size of English-centered datasets from training subword vocabulary, the SentencePiece model is trained by data aggregated from monolingual resources rather than paired resources.
We use Adam \cite{adam} optimizer with the inverse square root schedule in all cases and set different learning rates for different datasets.
For fair comparisons, we not only reproduce the Transformer \cite{attention} as Baseline but also reproduce the work of \citet{liu-2021} in all cases, which is denoted by Residual.
We always adopt the same hyperparameters setting to prior works and train corresponding subword vocabulary via details described by these works.
Specifically, as for the settings for cases of IWSLT, Triangle and Square, we follow the setting of \citet{liu-2021}; as for models trained by TED talks and OPUS-100, we follow the setting of \citet{philip-2020} and \citet{zhang-2020}, respectively.

\begin{table*}[!ht]
  \centering
  \resizebox{0.8\textwidth}{!}{
    \begin{tabular}{lcccccccc}
    \hline
   Supervised: &
   \multicolumn{2}{c}{IWSLT} &
   \multicolumn{1}{c}{Triangle} &
   \multicolumn{1}{c}{Square} &
   \multicolumn{2}{c}{TED} &
   \multicolumn{2}{c}{OPUS-100}  \\
    \hline
    Method & 
    en$\rightarrow$ & $\rightarrow$en &
    sup. &
    sup. &
    en$\rightarrow$ & $\rightarrow$en &
    en$\rightarrow$ & $\rightarrow$en \\
    \hline
    Baseline & 31.51 & 32.93 & 25.75 & 32.04 & 24.23 & 28.92 & 19.50 & 27.60 \\
    \citet{philip-2020} & \underline{ } & \underline{ } & \underline{ } & \underline{ } & 24.85 & \textbf{31.21} & \underline{ } & \underline{ } \\
    \citet{zhang-2020} & \underline{ } & \underline{ } & \underline{ } & \underline{ } & \underline{ } & \underline{ } & 21.39 & 27.50 \\
    Residual & 31.24 & 32.65 & 26.25 & 31.85 & 22.80 & 28.19 & 20.38 & 26.67\\
    SD & 31.63 & 32.51 & 26.50 & 31.97 & 23.94 & 28.33 & 23.60 & 28.01\\
    FCLL & \textbf{31.76} & \textbf{33.00} & \textbf{26.91} & \textbf{32.14} & \textbf{25.32} & 28.13 & \textbf{26.17} & \textbf{29.33}\\
    \hline
    \end{tabular}
  }
  \caption{
  Averaged BLEU scores on supervised directions.
  en$\rightarrow$ denotes translating from en ($l_c$) to $\mathbb{L'}$ and $\rightarrow$en denotes translating to en from $\mathbb{L'}$; sup. indicates supervised directions in non-centered cases.
  \emph{Residual} follows \citet{liu-2021} to modify residual connection.
}
  \label{tab:sup}
\end{table*}

\begin{table*}[!ht]
  \centering
  \resizebox{\textwidth}{!}{
    \begin{tabular}{lcccccccccccccc}
    \hline
    Zero-Shot: &
    \multicolumn{2}{c}{IWSLT} &
    \multicolumn{2}{c}{Triangle} &
    \multicolumn{2}{c}{Square} &
    \multicolumn{2}{c}{TED} &
    \multicolumn{4}{c}{OPUS-100}  \\
    \hline
    Method & 
    Z.S. & O.R. &
    Z.S. & O.R. &
    Z.S. & O.R. &
    Z.S. & O.R. &
    Z.S. & O.R. & F.T. & O.R.\\
    \hline
    Baseline & 16.97 & 13.95 & 1.97 & 93.68 & 31.18 & 0.74 & 10.66 & 4.16 & 3.97 & 63.96 & 10.11 & 13.92\\
    \citet{philip-2020} & \underline{ } & \underline{ } & \underline{ } & \underline{ } & \underline{ } & \underline{ } & 12.94 & \underline{ } & \underline{ } & \underline{ } & \underline{ } & \underline{ } \\
    \citet{zhang-2020} & \underline{ } & \underline{ } & \underline{ } & \underline{ } & \underline{ } & \underline{ } & \underline{ } & \underline{ } & 4.02 & 54.57 & 11.98 & \underline{ } \\
    Residual & 20.37 & 1.80 & 16.60 & 4.95 & 30.30 & 0.77 & 12.54 & 3.85 & 5.14 & 38.54 & 11.38 & 18.30 \\
    SD & \textbf{21.35} & 2.03 & 19.07 & 0.92 & 31.26 & 0.75 & 13.03 & 3.94 & 4.87 & 44.07 & 12.95 & 12.54 \\
    FCLL & 21.15 & 2.05 & \textbf{20.56} & 0.13 & \textbf{31.49} & 0.74 & \textbf{14.14} & 3.74 & \textbf{6.31} & 34.46 & \textbf{13.65} & 11.09 \\
    \hline
    \end{tabular}
  }
  \caption{
  Averaged BLEU scores on zero-shot directions.
  Z.S. column indicates results of zero-shot translation; O.R. denotes the off-target ratio measured by \%; F.T. indicates results after fine-tuning, we follow \citet{zhang-2020} to fine-tune 6 languages existing in zero-shot testing.
}

  \label{tab:zero}
\end{table*}
We experimented with IWSLT, Triangle, and Square five runs with different random seeds [1,2,3,4,5], to compute the variance for verifying the instability caused by parameters, and other experiments are trained with seed 1.
To evaluate results of all experiments, we translate the official test set with beam size 4, and evaluate the translation results by sacreBLEU \cite{bleu, sacrebleu}.
We also employ the langdetect\footnote{The tool is not accurate, so, it is just for observing general tendency. (\url{https://github.com/Mimino666/langdetect)}}, which can identify the language of one sentence, to count the off-Target ratio, namely how many sentences are not translated to the correct language.
We list detailed experimental settings in Appendix \ref{appendix:parameters}.

\subsection{Results}\label{section:result}
As described in Table \ref{tab:sup}, our proposed methods achieve small improvements measured by averaged BLEU scores on supervised directions of IWSLT (+0.25/+0.07), Triangle (+1.16), Square (+0.1), TED (+1.09/-0.79), and OPUS-100 (+6.67/+1.73) compared to Baseline.
\citet{liu-2021} speculated that the basic Transformer would overfit more on the supervised direction, and the improvement of zero-shot could hurt supervised translation \cite{gu-2019, zhang-2020, liu-2021}.
The performance of Residual \cite{liu-2021} degenerated in TED (-1.43/-0.73) and OPUS-100 (+0.88/-0.93), since the model weakened LS information by trading the generalization ability for zero-shot translation.
However, our proposed methods benefit from the additional improvement of decoding the target language by LS modeling \cite{sachan-2018, philip-2020}.
This improvement can counteract the insufficiency of tying artificial language tokens to instruct translation \cite{arivazhagan-2019}.
The results of SD can empirically prove the positive impact of CLL, since the performance of SD, which only constructs one CLL, is always between FCLL and Residual on supervised directions.
Moreover, the performance of FCLL shows a marked difference (+1.09/-3.08) from the work of \citet{philip-2020}.
We speculate that the reason is lacking LS structure of $l_c$ and benefiting from the mixture of shared and LS information in CLL.
This hypothesis can explain the stable improvements of CLL on Triangle (+1.16) and Square (+0.1) where no $l_c$ exited in training data.
We conduct ablation experiments to show the mechanism of CLL in Section \ref{section:mechanism}.

Table \ref{tab:zero} demonstrates that our methods always give the best scores on zero-shot translations in our experiments.
Based on the gain of zero-shot and gain of en$\rightarrow$ (Table \ref{tab:sup}), CLL always positively impacts non-centered languages.
In IWSLT, SD performs better than FCLL (+0.2) and performs near FCLL in other cases.
It indicates that stacking LS structures is not always optimal for improving zero-shot translation.
It also proves combining tweaking encoding information and improving decoding information would be effective for zero-shot translation.
In Triangle, our methods perform stably in the extreme data condition where Baseline totally failed.
In Square, all cases have similar performances since these languages do not have strict dependence.
Results in TED and OPUS-100 show that our methods also run well in the large-scale dataset.
Moreover, we follow \citet{zhang-2020} to fine tune the model by back-translation \cite{gu-2019} for 6 languages of zero-shot testing, and FCLL achieves a gain of +3.54 BLEU scores to Baseline.
These two points show the proposed CLL is orthogonal with other methods excluding LS modeling.

\begin{table}[t]
\centering
\resizebox{0.47\textwidth}{!}{
\begin{tabular}{lcccccc}
\hline
 & \multicolumn{3}{c}{\textbf{Zero-Shot}} & \multicolumn{3}{c}{\textbf{Supervised}} \\
\hline
 & Baseline & Residual & SD & Baseline & Residual & SD \\
\hline
(1) & 14.31 & 15.06 & \textbf{16.55} & 20.80 & 20.17 & \textbf{21.97} \\
(2) & 15.08 & 16.45 & \textbf{17.01} & \textbf{24.60} & 24.38 & 24.24 \\
\hline
\end{tabular}}
\caption{Averaged BLEU scores of integrating de. Row (1) and Row (2) shows results in
$\mathbb{L}_{iwslt} \rightarrow$ de and de $\rightarrow \mathbb{L}_{iwslt}$, respectively.}
\label{tab:integrate}
\end{table}

We further noticed that, in Table \ref{tab:sup}, the performances of all models on zero-shot directions in Square are comparable with each other, and our methods performed stably in Triangle where Baseline is totally failed.
The stability of Square case shows the key to improving zero-shot translation is not only large training data \cite{Mattoni-2017}, but also the balance of training \cite{shao-2022}.
The results of Triangle prove CLL is stable in zero-shot translation since it would not be influenced by different data conditions.
This feature ensured the effective utilization of shared information.
This feature can be proved by the value of off-target rate in Table \ref{tab:zero}.
Given the cost of establishing consistent semantic representation in shared information, confusion about different linguistic features is an inevitable result because the shared information container leads to coupling supervised translation pairs both in theory and practice, however, our proposed methods are always at a relatively lower rate.

\subsection{Integrating a new language by few data}
The ability to integrate a new language by few data is crucial for low-resource languages when extending a trained MNMT model.
To verify this ability of CLL, we fine tune trained SD in IWSLT and extend it to German (de) using bilingual language pairs (en $\leftrightarrow$ de) with 15K sentences per direction, we also fine tune Baseline and Residual as comparison.
We follow \citet{liu-2021} to set hyperparameters and update subword vocabulary, as described in Appendix \ref{appendix:parameters}.
Table \ref{tab:integrate} shows that SD performs better on zero-shot translation, which indicates CLL contributes the cross-lingual knowledge transfer, which indicates that our method is flexible in incorporating low-resource languages.

\section{Discussion and Analyses}
\begin{table}[t]
\centering
\resizebox{0.47\textwidth}{!}{
\begin{tabular}{lcccccc}
\hline
 & \multicolumn{2}{c}{\textbf{IWSLT}} & \multicolumn{2}{c}{\textbf{Triangle}} & \multicolumn{2}{c}{\textbf{Square}} \\
\hline
 & sup. & zero. & sup. & zero.
 & sup. & zero.\\
\hline
Baseline & 0.021 & 5.280 & 0.220 & 0.210 & \textbf{0.001} & 0.016\\
Residual & 0.067 & 0.270 & \textbf{0.004} & 0.900 & 0.003 & 0.055\\
SD & \textbf{0.012} & \textbf{0.051} &  0.018 & 0.900 & 0.002 & \textbf{0.001}\\
FCLL & 0.025 & 0.074 & 0.018 & \textbf{0.140} & 0.004 & 0.014\\
\hline
\end{tabular}}
\caption{Variance computed from averaged BLEU scores among five runs in IWSLT, Triangle, and Square with different random seeds. sup. and zero. indicate supervised translation and zero-shot translation, respectively. Smaller variance means a more stable result.
}
\label{table:variance}
\end{table}
\subsection{Instability of Zero-Shot Translation}\label{section:instability}
In this paper, we describe the instability from two related perspectives: 1) Instability of training; 2) Instability of data conditions.
For the first point, Table \ref{table:variance} shows the variance for different models counted from five experiments with different seeds for initialization.
The small value of variance on supervised translation among the four models shows that supervised training always is a relatively stable process.
However, the training process of zero-shot translation is sensitive to initial parameters \cite{gu-2019}, since the variance of zero-shot translation is always higher than the variance of supervised translation.
Our methods always achieved the lowest variance on zero-shot translation.
For the second point, Table \ref{tab:zero} shows that Baseline has completely lost its ability of zero-shot translation in Triangle, although the amount of training samples of Triangle is relatively higher than IWSLT that can result in a good performance on zero-shot translation.
On the other hand, Square performs excellently on zero-shot translation and its performance is even closing to the performance of supervised translation (Table \ref{tab:zero}), although it is non-centered data condition as same as Triangle and it is not completely interconnecting all languages \cite{fan,freitag-2020}.
These comparisons proved that the data condition impacts the learning of zero-shot translation.

We further notice that Baseline has a high variance in IWSLT yet a small variance in Triangle.
We speculate that the strict dependence of non-centered languages caused instability, and the degree of dependence influences the expression of instability.
Specifically, Baseline tends to build cross-linguistic representations in IWSLT, but the strict dependence would couple representations of non-centered languages to the central language to lead to a high off-target ratio in testing zero-shot translation (Table \ref{tab:zero}).
And the higher variance of Baseline in IWSLT means that the model may find a special set of initial parameters to escape from the negative influence of strict dependence.
Moreover, the small variance of Baseline in Triangle means that the model completely cannot find a special set among the five times experiments, since Triangle has the most severe dependence of non-centered languages.

\begin{table}[t]
\centering
\resizebox{0.47\textwidth}{!}{
\begin{tabular}{clccc}
\hline
& \textbf{Condition}& it$\rightarrow$nl& ro$\rightarrow$it& nl$\rightarrow$ro\\
\hline
 (1)& Baseline & 18.69 & 16.43 & 14.13 \\
 (2)& (1)+additional pairs & 23.27 & 22.17 & 21.61 \\
 (3)& (2)+reduce data & 22.35 & 21.31 & 20.96 \\
 & SD & 22.13 & 20.27 & 20.42 \\
\hline
\end{tabular}}
\caption{Variation of different conditions in IWSLT. SD is the performance under original setting.}
\label{tab:valid}
\end{table}

To prove our speculation, we create two artificial setups based on IWSLT to re-train Baseline and show results measured by BLEU in Table \ref{tab:valid}.
Specifically, for Row (2), we append three language pairs (it $\rightarrow$ ro, ro $\rightarrow$ nl, nl $\rightarrow$ it) with 30k sentences per pair to balance the dependence \cite{rios-2020}; for Row (3), we sample a random subset of 90K sentences from training data of IWSLT of 145K sentences per translation direction and we append additional pairs as (2).
These substantial gains (up to +7.48) of Row (2) in Table \ref{tab:valid} proved our viewpoint that data conditions impact the performance of zero-shot translation. 
Once the model disentangled the strict dependence by appending additional pairs, the model would achieve considerable gains (up to +6.83), although the training samples have been reduced to be smaller than the original setting shown by Row (1).
Moreover, the performance of SD is comparable to these artificial cases.

So far we can conclude that the strict dependence of non-centered languages closely influences the zero-shot translation.
And our motivation for disentangling the dependence by improving the weak representations of non-centered languages is effective.
We will discuss the mechanism of CLL in Section \ref{section:mechanism}.

\begin{figure*}[ht]
    \centering
    \begin{subfigure}[b]{0.32\textwidth}
        \centering
        \includegraphics[width=\textwidth]{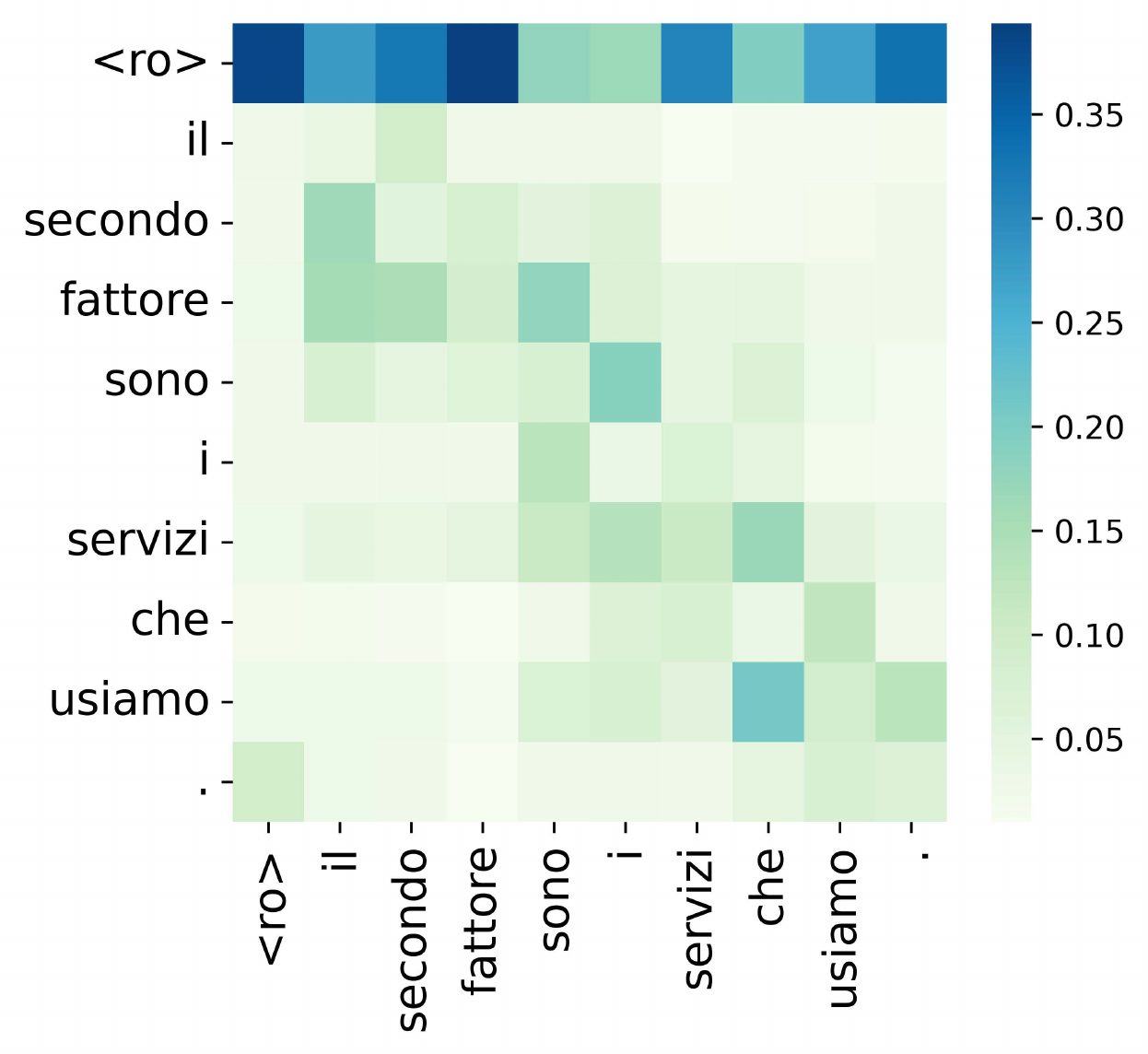}
        \caption{CLL}
        \label{figure31}
    \end{subfigure}
    \hfill
    \begin{subfigure}[b]{0.32\textwidth}
        \centering
        \includegraphics[width=\textwidth]{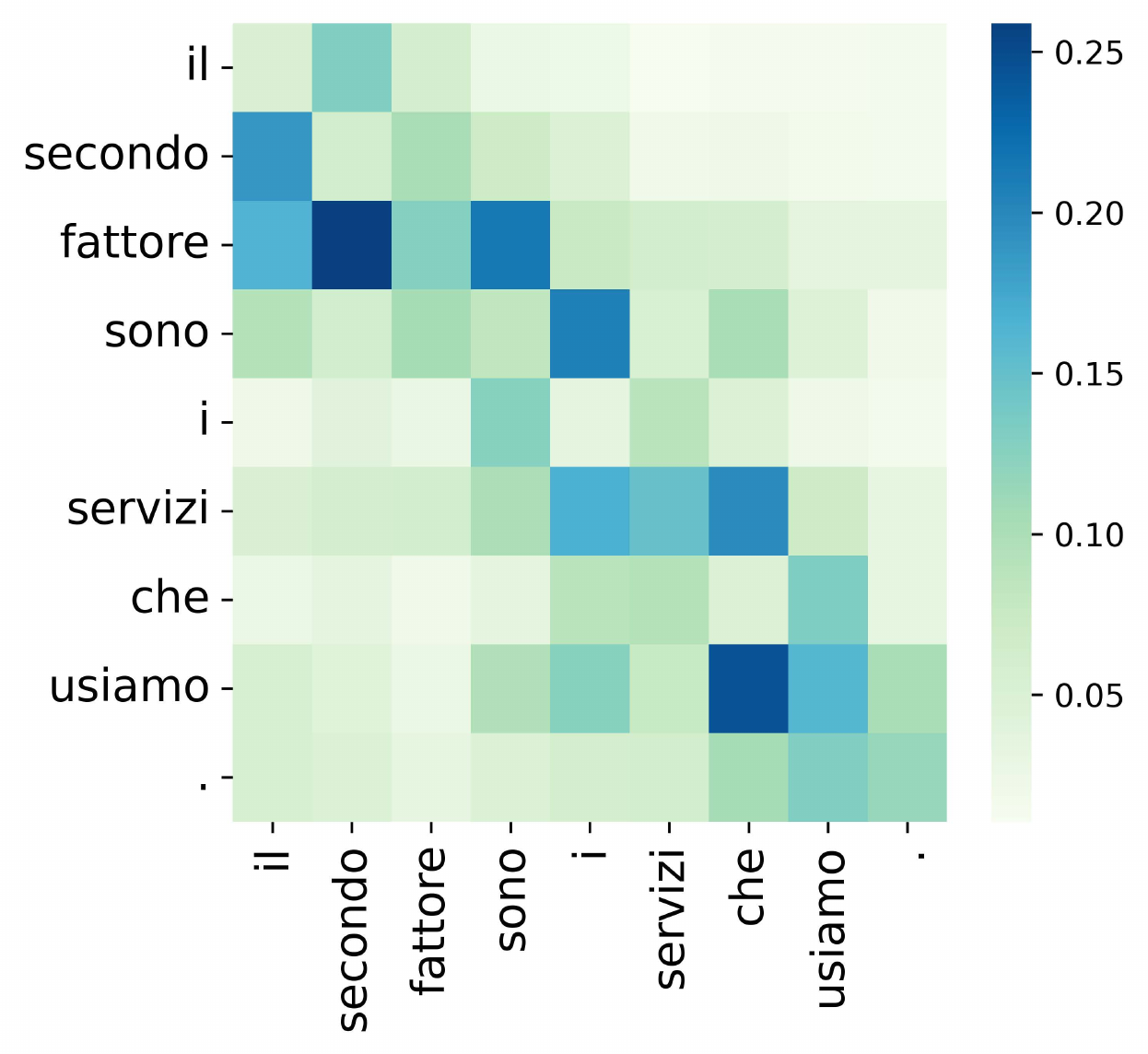}
        \caption{CLL w/o token}
        \label{figure32}
    \end{subfigure}
    \hfill
    \begin{subfigure}[b]{0.32\textwidth}
        \centering
        \includegraphics[width=\textwidth]{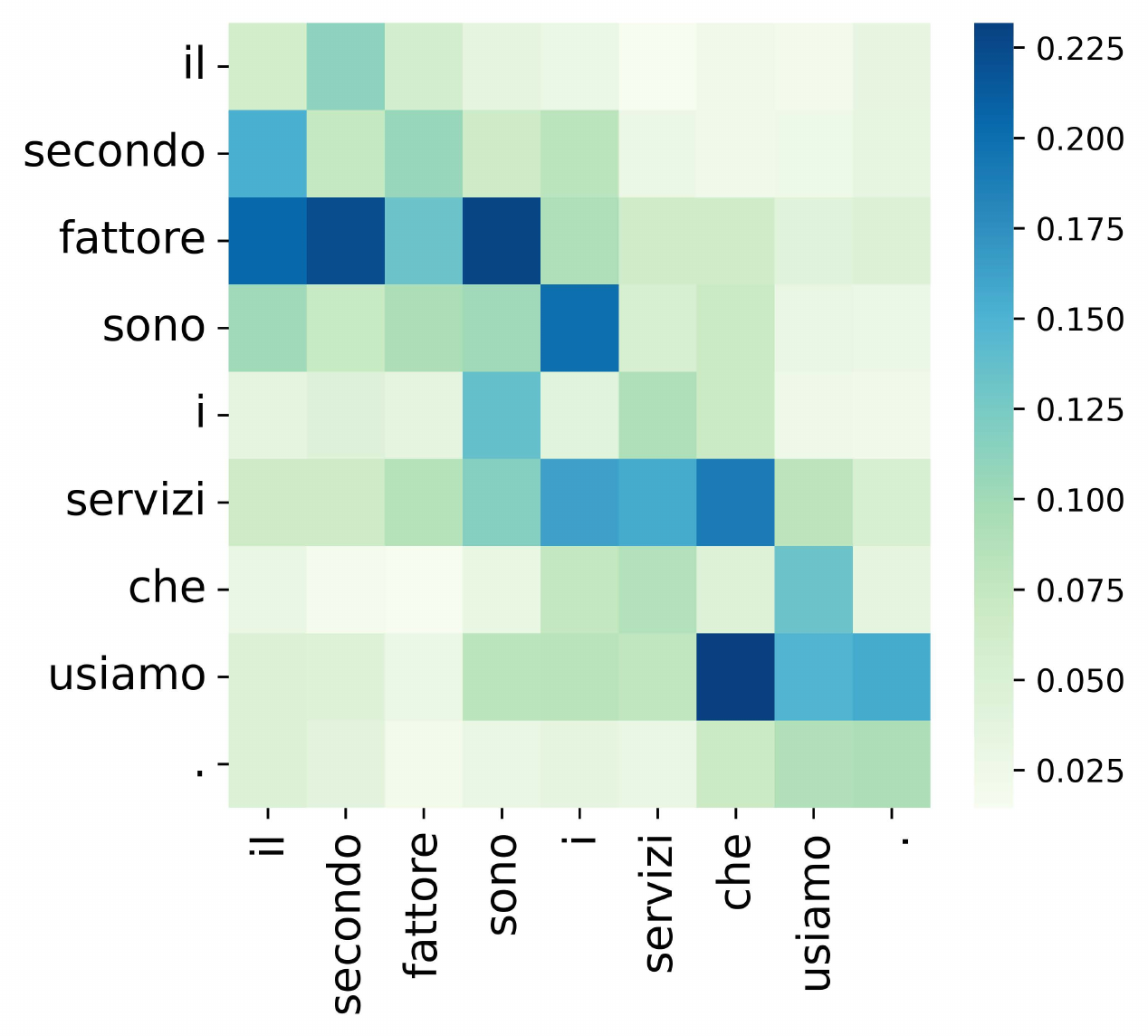}
        \caption{CLL omitted token}
        \label{figure33}
    \end{subfigure}
\caption{Maps of Self-Attention in which translating one sentence of ro $\rightarrow$ it.}
\label{figure3}
\end{figure*}
\subsection{Translation Instructions}\label{section:instruction}
We re-train our models in IWSLT without language tokens \cite{Johnson}, and Table \ref{tab6} shows the result.
First, slight performance gains were observed on supervised directions in FCLL and SD.
Figure \ref{figure3} is a heat map showing self-attention weights of FCLL with and without language tokens.
Figure \ref{figure31} shows one possibility is artificial language tokens \cite{Johnson} might disturb the semantic representation for actual words, since the language token <ro> dominated in self-attention weights.
Figure \ref{figure32} shows the distribution of actual words weights by training without language tokens.
Figure \ref{figure33} presents the attention weights when omitting <ro> in testing the model trained with language tokens, and we observed a similar tendency with the plot in Figure \ref{figure32}.

Second, Table \ref{tab6} shows our methods stably maintain cross-linguistic representation although no language tokens were inserted to instruct translation directions both for supervised and zero-shot directions.
On the contrary, other methods completely lost their ability of zero-shot translation.
These analyses indicate that CLL has a strong capability to instruct multilingual translation.
\begin{table}[t]
\centering
\resizebox{0.47\textwidth}{!}{
\begin{tabular}{lcccc}
\hline
\textbf{Method}& \textbf{Supervised}& \textbf{Zero-Shot}& \textbf{Off(\%)}\\
\hline
 Residual & 32.57 & 20.74 & 1.67 \\
 FCLL & 32.95 & 21.00 & 1.52 \\
 SD & 32.48 & 21.16 & 1.97 \\
\hline
 Residual w/o t & 23.72 & 0.56 & \underline{ } \\
 FCLL w/o t & 32.88 & 20.85 & 1.35 \\  
 SD w/o t & 32.62 & 19.46 & 2.10 \\
\hline
\end{tabular}}
\caption{Averaged BLEU scores of models training without language tokens (w/o t) in IWSLT.}
\label{tab6}
\end{table}
\begin{table}[t]
\centering
\resizebox{0.47\textwidth}{!}{
\begin{tabular}{lcccccc}
\hline
 & \multicolumn{2}{c}{it$\rightarrow$} & \multicolumn{2}{c}{ro$\rightarrow$} & \multicolumn{2}{c}{nl$\rightarrow$}\\
\hline
 & ro & nl & it & nl & it & ro \\
\hline
FCLL  &21.14 &21.92 &20.43 &22.12 &19.45 &20.98 \\
Omitted  & -0.16 & -0.08 & -0.37 & -0.21 & -0.87 & -0.74 \\
\hline
SD  &20.88 &22.13 &20.27 &22.75 &20.42 &20.51 \\
Omitted  & -1.79 & -1.40 & -2.13 & -1.68 & -2.41 & -2.59 \\
\hline
\end{tabular}}
\caption{Variation of BLEU scores after omitting language tokens in testing.}
\label{tab7}
\end{table}

\subsection{Full Layers vs. Single Layer}\label{section:number}
To verify whether the number of CLL affects the performance of MNMT model, we translate the test set omitting artificial language tokens by trained FCLL and SD in IWSLT.
Table \ref{tab7} demonstrates the model with more CLL layers has stronger robustness since scores of FCLL degenerated less than SD.
To further investigate the effect of the number of CLL layers, we re-train Transformer models with different numbers of CLL layers based on IWSLT in two cases. 
Specifically, in case (1), we modify the residual connection of models as same as the operation of SD, but we do not modify any architecture in encoder of Transformer in case (2).
Then, we follow the idea of \citet{liu-2021} to remove the CLL layers in the decoder from the top-most and bottom-most positions until the configuration in which only a single CLL layer is preserved in the middle-position decoder among all decoders of these models.
\begin{figure}[t]
    \centering
    \includegraphics[width=0.47\textwidth]{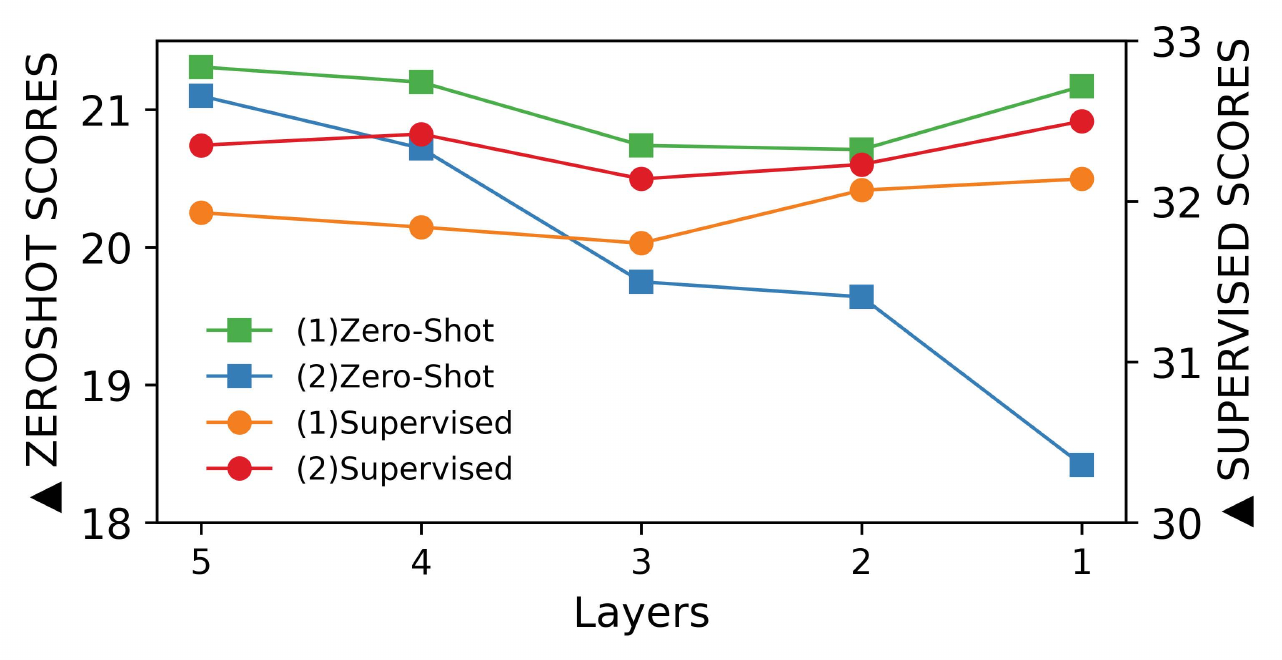}
    \caption{Variation of BLEU scores in which training on different CLL layers. (1) modified the residual connection as SD; (2) did not modify it.}
    \label{figure4}
\end{figure}
\begin{figure*}[t]
    \centering
    \begin{subfigure}[b]{0.31\textwidth}
        \centering
        \includegraphics[width=\textwidth]{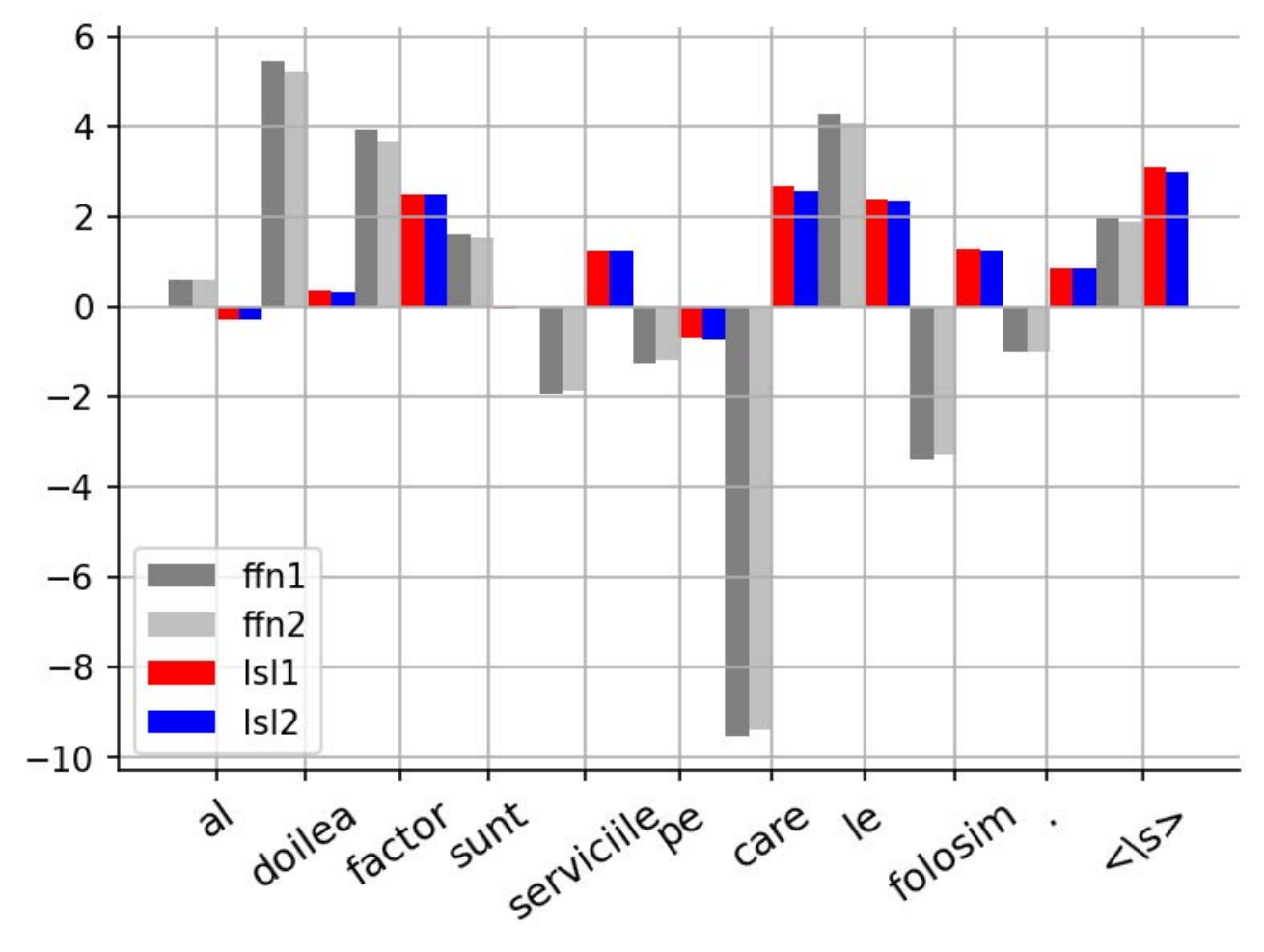}
        \caption{IWSLT:it $\rightarrow$ ro}
        \label{figure5.1}
    \end{subfigure}
    \hfill
    \begin{subfigure}[b]{0.31\textwidth}
        \centering
        \includegraphics[width=\textwidth]{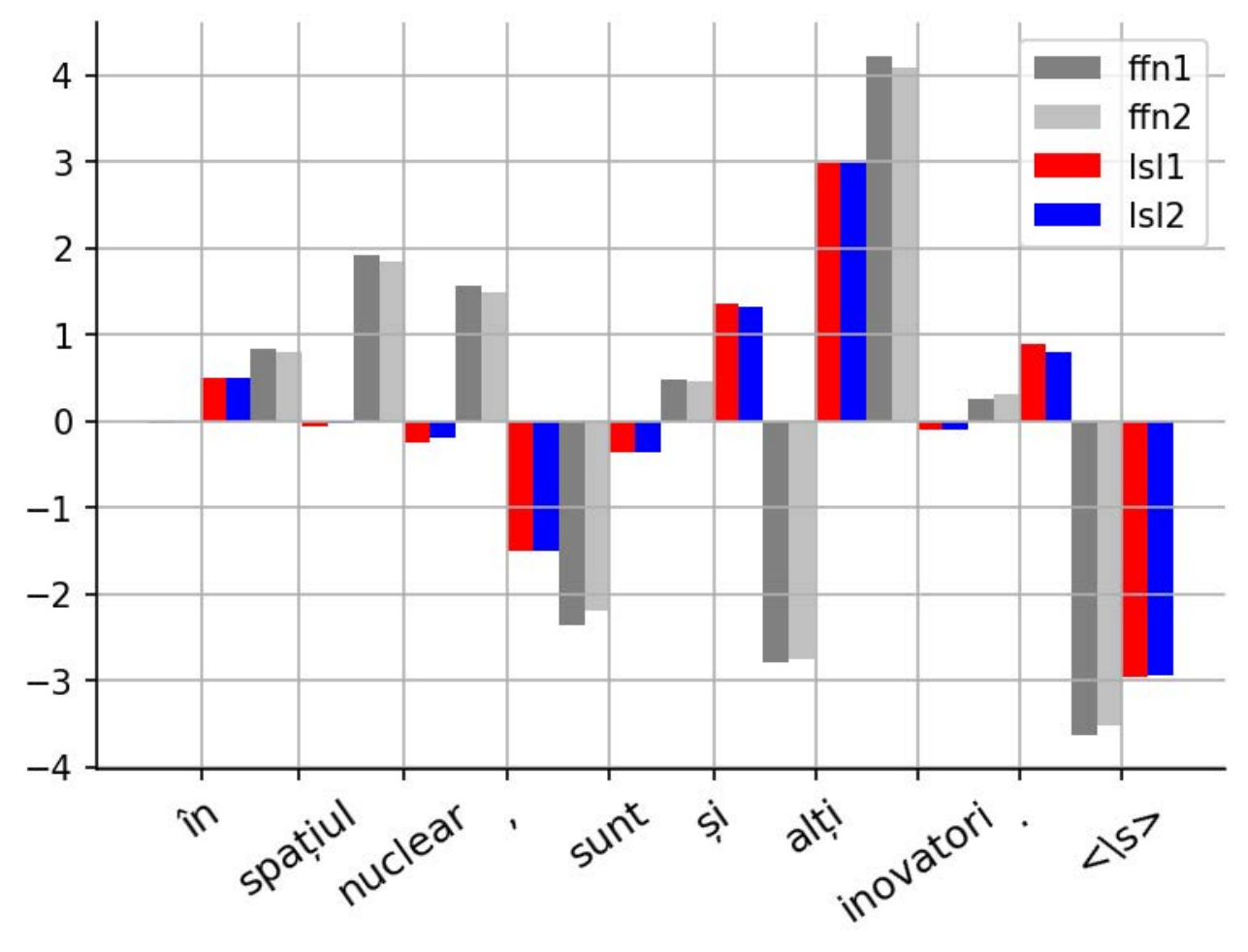}
        \caption{IWSLT:en $\rightarrow$ ro }
        \label{figure5.2}
    \end{subfigure}
    \hfill
    \begin{subfigure}[b]{0.31\textwidth}
        \centering
        \includegraphics[width=\textwidth]{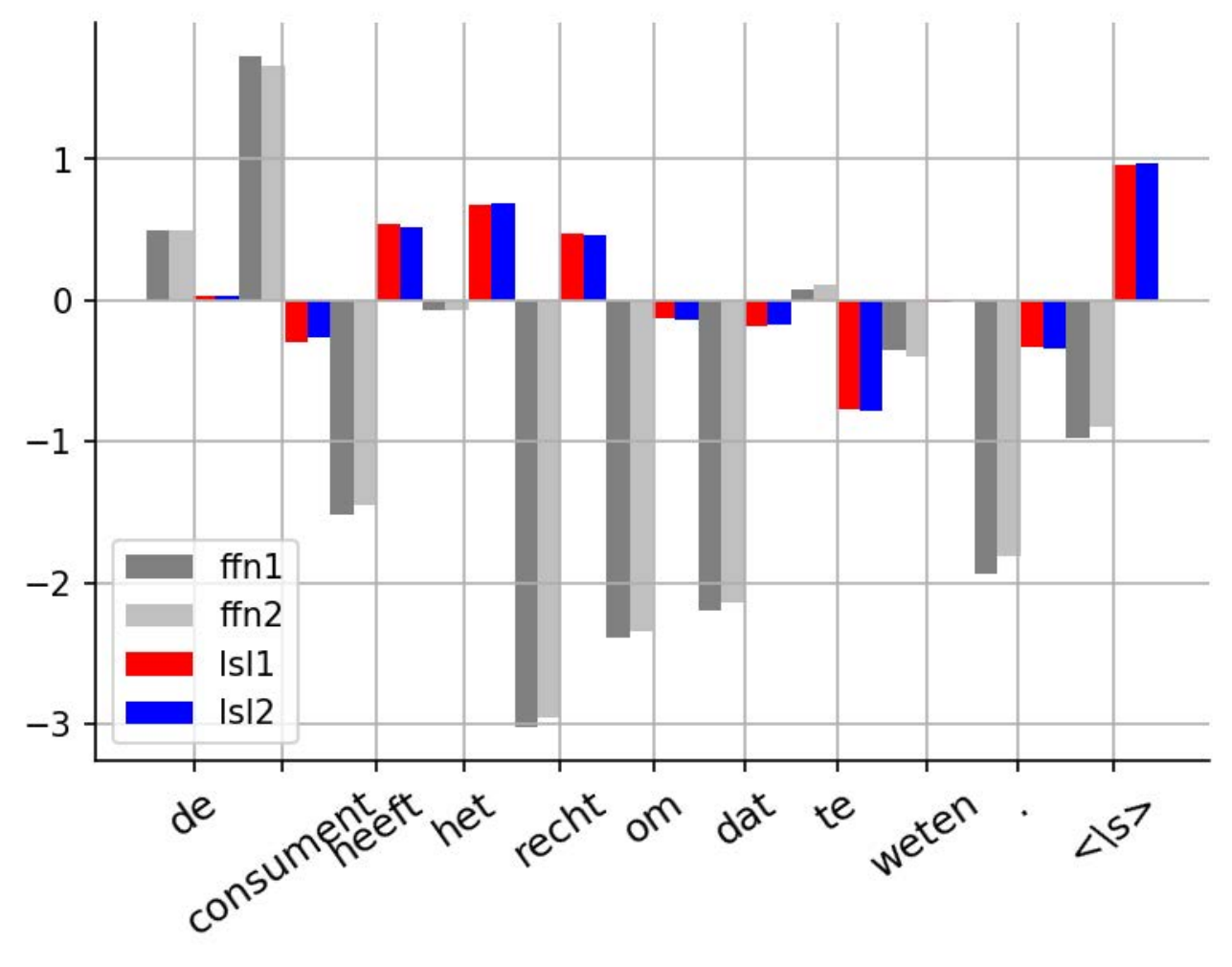}
        \caption{Triangle:de $\rightarrow$ nl }
        \label{figure5.3}
    \end{subfigure}
    \caption{Visualization of layer attributions. ffn indicates FFN, lsl means LSLs, 1 or 2 means it is 1st or 2nd fully connected neural network of this component. A higher absolute value indicates more contribution for result.}
    \label{figure5}
\end{figure*}

Figure \ref{figure4} shows that the zero-shot performance of models in (2) degenerated with the reduction of the number of CLL layers, although the supervised performance always kept in the same magnitude.
It proves that the increase in the number of CLL layers has a positive impact on the zero-shot translation.
However, almost no clear variations were observed in (1) of Figure \ref{figure4}.
One possibility of the lower supervised performance of (1) when compared with (2) in Figure \ref{figure4} is the weaken language specific information in the encoder by removing the residual connection \cite{liu-2021}.
Likewise, the zero-shot performance of (1) is not sensitive to the variation of the number of CLL layers since weakening the capacity of the encoder could partially offset CLL's gains in decoder.
We conclude that the architecture of SD is relatively-optimal in small-scale dataset because it is lightweight yet comparable with FCLL, and FCLL is more stable where data condition is complex or large.

\begin{table}[t]
    \centering
    \resizebox{0.47\textwidth}{!}{
    \begin{tabular}{lcccccc}
    \hline
    & \multicolumn{3}{c}{\textbf{supervised}} & \multicolumn{3}{c}{\textbf{zero-shot}} \\
    & it$\rightarrow$ & nl$\rightarrow$ & de$\rightarrow$ & it$\rightarrow$ & nl$\rightarrow$ & de$\rightarrow$\\
    \hline
    (1) & 26.80 & 26.28 & 26.00 & 18.44 & 19.80 & 16.36 \\
    (2) & 25.19 & 25.77 & 25.56 & 0.64 & 0.75 & 1.02 \\
    (3) & 24.85 & 24.46 & 25.49 &\underline{ } &\underline{ } &\underline{ } \\
    \hline
    \end{tabular}
    }
    \caption{Averaged BLEU scores of ablation study. Row (1) shows results in Triangle; Row (2) shows results after ablation; Row (3) means to calculate scores of zero-shot translation by treating the supervised translation results as reference data.}
    \label{tab:ablate}
\end{table}

\begin{table}[t]
\centering
    \resizebox{0.47\textwidth}{!}{
    \begin{tabular}{cc}
         \hline
          it$\rightarrow$de & Ablated LSL of de from CLL\\
         \hline
         Input: & \makecell[l]{<de> la quarta priorità concerne l’attenzione\\ 
         che occorre prestare ai nuovi rischi.}\\
         \hline
         \makecell[c]{Expected \\
         Output:} & 
         \makecell[l]{die vierte priorität gilt den neuen risiken.}\\
         \hline
         \makecell[c]{Actual \\
         Output:} &
         \makecell[l]{de vierde prioriteit is de aandacht die moet \\
         worden besteed aan nieuwe risico ’s.}\\
    \hline
    \end{tabular}
    }
    \caption{Ablated testing SD trained in Triangle.
    The output of the model rolls back to nl (Dutch, the supervised direction).}
    \label{tab:case}
\end{table}
\subsection{Disentangling Coupled Representation}\label{section:mechanism}
\paragraph{Ablation Study}
To investigate the significance of CLL, we ablate LSL from CLL of trained SD in Triangle, namely only use FFN, to re-translate the test set.
Row (2) of Table~\ref{tab:ablate} show the degeneration of supervised translation is not significant, but completely losing zero-shot translation capability.
However, we observed that the zero-shot translation rollbacks to supervised directions after ablation via analyzing failure cases.
As shown in Table \ref{tab:case}, the zero-shot translation of it $\rightarrow$ de will be biased to it $\rightarrow$ nl due to ablating the layer of de\footnote{we report more examples in Appendix \ref{appendix:case} in which including long and short sentences in different cases.}.
Thus, we calculated the BLEU scores of zero-shot translation by treating the test set of it $\rightarrow$ nl as reference data in testing Row (3) of Table \ref{tab:ablate}.
The slight degeneration of Row (3) strongly proved that FFN has built a consistent semantic representation which has been coupled to supervised directions.

\paragraph{Layer Attributions}
The layer attribution\footnote{We employ Captum  (\url{https://github.com/pytorch/captum}) for computing attributions.} can quantify the contributions of one component by integrated gradients \cite{captum}.
We designed 3 scenarios to observe these attributions in details:
a) The zero-shot translation based on centered case;
b) The supervised translation based on centered case;
c) The zero-shot translation based on non-centered case.
Figure \ref{figure5} demonstrates:
1) FFN always plays the main role in translation;
2) Generally, the contributions of CLL are on the contrary of FFN in LS words, but they have similar contributions in common words, especially the punctuation.

These results proved our viewpoint in Section \ref{section:instability} again.
Specifically, the shared representations built in FFN potentially enable cross-linguistic transferring, but the strict dependence of non-centered languages would hamper freely transferring since cross-linguistic information is coupled with supervised translation directions.
Therefore, the significance of LSL in CLL practically is to provide independent LS information to disentangle the coupled representation, namely counteract the negative influence of the dependence, to present a correct LS representation in decoding.

\section{Conclusion}
In this work, we supplement the theory of zero-shot translation with the strict dependence of non-centered languages, and we describe the instability of zero-shot translation.
To counteract the influence of the dependence, we proposed a simple yet effective method that employs LS modeling by adapting to non-centered languages.
Our analysis based on layer attribution demonstrated that LS information is conducive to disentangling the coupled model representation.
Our experiments on various datasets and different data conditions show that our proposed method outperforms in performance and complexity.

\section{Ethical Considerations}
The potential ethical risk of our work is the usage of multilingual datasets including IWSLT, Europarl, TED talks and OPUS-100, since these datasets might contain social biases, especially in the Europarl, in which predominant European languages might constitute stereotypes.
Those biases would be represented in the trained model and could be amplified by integrating one new language out of trained language families since no special treatment is performed to mitigate the biases.
Generally, this method can be landed in the industry under sufficient anti-prejudice measures.

\section*{Acknowledgements}
This work was in part supported by JSPS KAKENHI Grant Numbers 21H05054.

\bibliography{anthology,custom}
\bibliographystyle{acl_natbib}

\clearpage
\appendix

\section{Detailed Settings}\label{appendix:parameters}
\paragraph{IWSLT \& Triangle \& Square}
We follow \citet{liu-2021} to set 5 encoder/decoder layers with 8 attention heads, embedding size of 512, inner size of 2048, dropout rate of 0.3, dropout rate of CLL layer of 0.3, maximum learning rate of 0.0005 and label smoothing rate of 0.1.
However, we decrease dropout rate to 0.1 and dropout rate of CLL layer to 0.2 in Square that is a bigger case than others.
The size of subword vocabulary is 40K for each case.
In training, we set the maximum batch size per GPU to 4,000 tokens, and train on 4 GPUs.
We train for 100K steps for IWSLT and Triangle, but train for 500K steps for Square.
We sample the supervised and zero-shot translation directions from the dev set of MMCR4NLP as the validation dataset in training.

\paragraph{TED talks}

We follow \citet{philip-2020} to set 6 encoder/decoder layers with 4 attention heads, embedding size of 512, inner size of 1024, dropout rate of CLL layer of 0.3, maximum learning rate of 0.0005 and label smoothing rate of 0.1.
However, we set the dropout rate to 0.2 to get better performances. 
The size of subword vocabulary is 70K.
In training, we set the maximum batch size per GPU to 4,000 tokens, and train on 4 GPUs.
We train for 90 epochs to ensure models convergent.
We only sample dev sets of supervised directions translating as the validation dataset in training.
We also follow \citet{philip-2020} to use mixed-precision \cite{ott-2018} in training.

\paragraph{OPUS-100}

We follow \citet{zhang-2020} to set 6 encoder/decoder layers with 8 attention heads, embedding size of 512, inner size of 2048, dropout rate to 0.1, dropout rate of CLL layer of 0.2, maximum learning rate of 0.0007 and label smoothing rate of 0.1.
We directly reuse their published subword vocabulary\footnote{https://github.com/bzhangGo/zero}.
In training, we set the maximum batch size per GPU to 6,000 tokens, and train on 8 GPUs\footnote{We use Fairseq command line of \emph{--update-freq 2} to simulate the efficiency of 8 GPUs by 4 GPUs.} for 500K steps.
We follow \citet{zhang-2020} to sample top 200 sentences in dev sets of supervised directions translating as the validation dataset in training.

In fine-tuning, we follow \citet{zhang-2020} to back-translate the training resource to get the pseudo resource, then we merge real and pseudo resources to train 4 epochs, and we update the pseudo training resource after each epoch in training.
We set 500 warm-up steps at the beginning of fine-tuning, reset the optimizer, and training with maximum learning rate of 0.0003.

\paragraph{Integrating de in IWSLT}
Based on the trained model in IWSLT, we learn a new SentencePiece model with 10K vocabulary size to acquire a dictionary for de.
Then we append the new dictionary to the end of the previously learned dictionary of IWSLT, meanwhile, we keep the order of the previous part unchanged.
Due to the increased number of unique tokens, we resize token embedding and initialize new vectors as the average of existing embedding perturbed by random noise.
When finetuning, we set the learning rate as the value at the end of the previous training, freeze parameters of CLL layers of existing languages, initialize parameters of CLL layers for de by averaging existing CLL layers, and include the original training data of IWSLT to prevent the shared information from tending to translate de.

\begin{table*}[t]
\section{Dataset Details}\label{appendix:dataset}
\centering
\resizebox{0.90\textwidth}{!}{
\begin{tabular}{lccc}
\hline
\makecell[l]{\textbf{Dataset} \\ case} & Languages & \makecell[c]{  \# zero-shot \\ directions} & \makecell[c]{  \# sent. \\ per direction} \\
\hline
\makecell[l]{\textbf{IWSLT} } & \{\underline{en}, it, ro, nl\} & 6 & 145K \\

\hdashline[5pt/2pt]
\makecell[l]{\textbf{Europarl} \\ Triangle} & \{\underline{\quad}, it, nl, de\} & 3 & 200K \\

\makecell[l]{\textbf{Europarl} \\ Square} & \{\underline{\quad}, fr, it, de, es\} & 4 & 1M \\

\hdashline[5pt/2pt]
\textbf{TED}& \makecell[c]{
\{
\underline{en}, ar, bg, de, es, fa, fr, he, hu, it, ja,\\
ko, nl, pl, pt-br, ro, ru, tr, vi, zh-cn
\}} & 342 & 140K \textasciitilde 210K \\

\hdashline[5pt/2pt]
\makecell[l]{\textbf{OPUS-100}} & \makecell[c]{
\{
\underline{en}, an, as, be, bg, bn, br, bs, ca, cs, cy,\\
da, de, el, es, fa, fr, fy, ga, gd, gl, gu, hi, \\
hr, hy, is, it, ku, li, lt, lv, mk, mr, nb, ne, \\
nl, nn, no, oc, or, pa, pl, ps, pt, ro, ru, sh, \\
si, sk, sl, sq, sr, sv, tg, uk, ur, wa, yi, az, \\
kk, ky, tk, tr, tt, ug, uz, dz, my, zh, et, fi, \\
hu, se, id, km, mg, ms, vi, ig, rw, xh, yo,  \\
zu, kn, ml, ta, te, eo, eu, ja, ko, ka, mn, th\}} & 30 & 2K \textasciitilde 1M \\
\hline
\end{tabular}}
\caption{Overview of datasets. The underline denotes the ${l}_c$, and the underline with blank represents non-centered condition, i.e. no English.}
\label{tab:dataset}
\end{table*}

\begin{figure*}[t]
\section{Distribution of Language-Specific Information Weights}\label{appendix:dis}
    \centering
    \includegraphics[width=\textwidth]{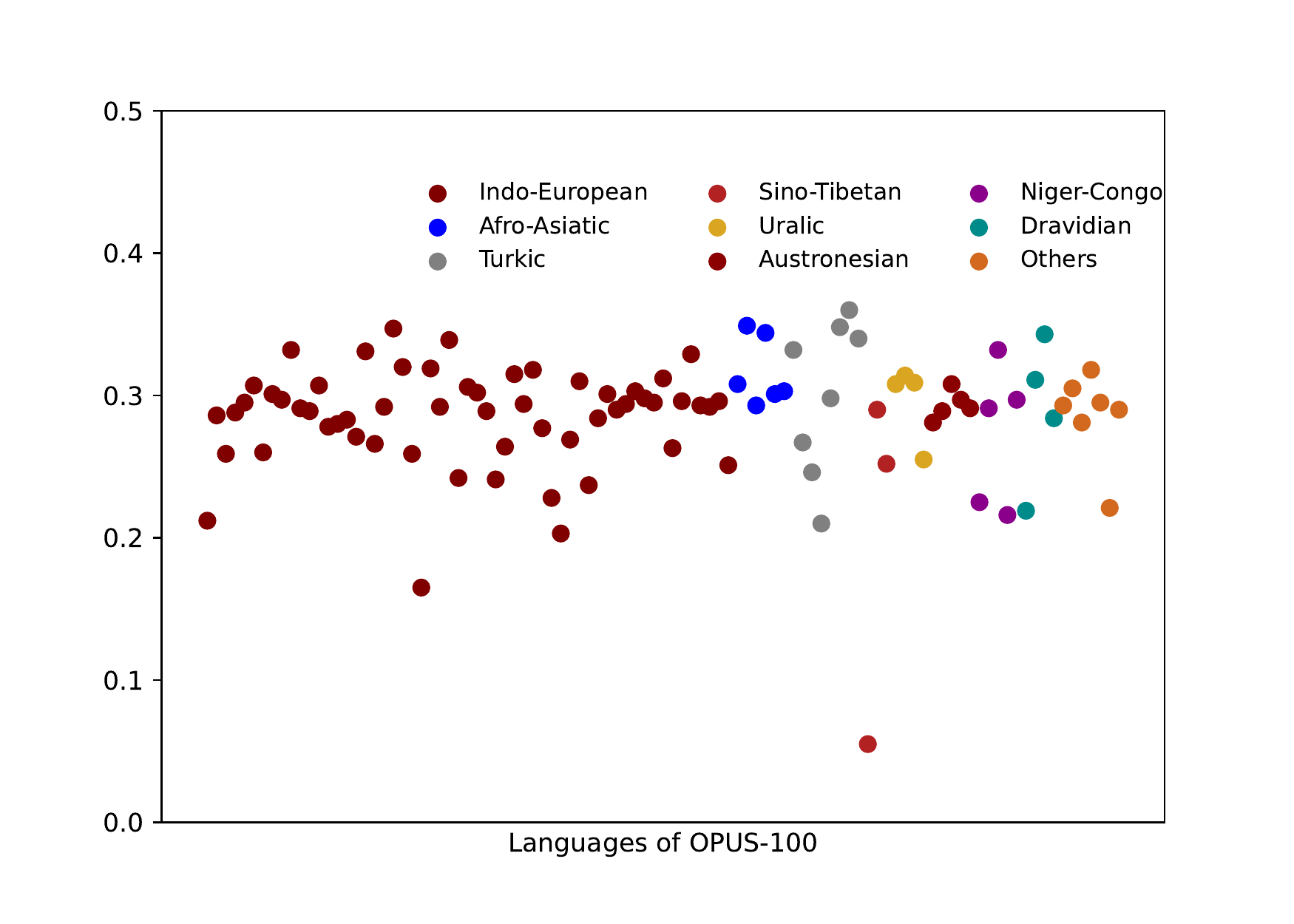}
    \caption{
    Averaged weights over all layers in FCLL model. The x-coordinate is sorted by languages showed in Table \ref{tab:dataset}. For languages with the same amount of training resources, languages from the same language family have relatively similar weights.
    }
    \label{figure6}
\end{figure*}

\begin{table*}[t]
\section{Translation Examples of Ablation}\label{appendix:case}
    \centering
    \resizebox{0.90\textwidth}{!}{
    \begin{tabular}{lcc}
    \hline
    Supervised: it$\rightarrow$nl & Ablated LSL of nl from CLL & Language\\
    \hline
    Input: & \makecell[l]{<nl> parlo adesso per esperienza personale: da anni nell'industria dell'aviazione\\ civile e con la commissione siamo infarciti di deregolamentazione, eppure,in re-\\lazione ai diritti aeroportuali, ci viene detto adesso che la risposta è la regolame-\\ntazione.} & it \\
    Expected Output: & \makecell[l]{ik spreek nu namens mijzelf: al vele jaren wordt ons nu binnen de burgerlucht-\\vaartindustrie en met de commissie een dieet voorgeschoteld van deregulatie en\\ toch, waar het gaat om luchthavenbelasting, wordt ons nu verteld dat regelgeving\\ het antwoord is.} & nl\\
    Actual Output: & \makecell[l]{ik spreek nu uit persoonlijke ervaring: al jaren in de burgerluchtvaartindustrie en\\ met de commissie zijn we gedwongen tot deregulering, maar wat de luchthaven-\\gelden betreft, wordt ons nu gezegd dat het antwoord de regelgeving is.} & nl \\
    \hline
    Input: & \makecell[l]{<nl> tutte le cose importanti sono già state dette.} & it\\
    Expected Output: & \makecell[l]{al het belangrijke is reeds gezegd.} & nl\\
    Actual Output: & \makecell[l]{al het belangrijke is reeds gezegd.} & nl\\
    \hline
    Supervised: de$\rightarrow$it & Ablated LSL of it from CLL & Language\\
    \hline
    Input: & \makecell[l]{<it> der verbraucher hat ein recht darauf, das zu wissen.} & de \\
    Expected Output: & \makecell[l]{il consumatore ha il diritto di saperlo.} & it \\
    Actual Output: & \makecell[l]{il consumatore ha il diritto di saperlo.} & it \\
    \hline
    Zero-Shot: it$\rightarrow$de & Ablated LSL of de from CLL & Language\\
    \hline
    Input: & \makecell[l]{<de> il recepimento di parte dell'acquis nel primo pilastro apre la strada alla co-\\munitarizzazione di questa politica e consente di adottare anche rimedi in relaz-\\ione alla nebulosa schengen, come amava chiamarla il mio predecessore.} & it\\
    Expected Output: & \makecell[l]{mit der teilweisen übernahme des acquis in den ersten pfeiler stehen uns nun alle\\ wege offen, diese politik zu vergemeinschaften und licht in la nébuleuse scheng-\\en zu bringen, wie es mein vorredner beschrieb.} & de\\
    Actual Output: & \makecell[l]{de omzetting van een deel van het acquis in de eerste pijler maakt de weg vrij vo-\\or de communautarisering van dit beleid en maakt het mogelijk dat er ook oplos-\\singen worden gevonden voor de nebulosa schengen, zoals mijn voorganger zei.} & nl\\
    \hline
    Input: & \makecell[l]{<de> la quarta priorità concerne l'attenzione che occorre prestare ai nuovi rischi.} & it \\
    Expected Output: & \makecell[l]{die vierte priorität gilt den neuen risiken.} & de\\
    Actual Output: & \makecell[l]{de vierde prioriteit is de aandacht die moet worden besteed aan nieuwe risico 's.} & nl\\   
    \hline
    Zero-Shot: nl$\rightarrow$it & Ablated LSL of it from CLL & Language\\
    \hline
    Input: & \makecell[l]{<it> die solidariteit en die noodzaak tot samenwerking geldt ook als zich in de\\ toekomst problemen voordoen, bijvoorbeeld bij interne migratiestromen.} & nl\\
    Expected Output: & \makecell[l]{questa sicurezza e la necessità di una collaborazione sono essi stessi potenziali\\ problemi futuri, ad esempio per quanto riguarda la migrazione interna.} & it\\
    Actual Output: & \makecell[l]{diese solidarität und die notwendigkeit der zusammenarbeit gelten auch in zu-\\kunft, z. b. in bezug auf die migrationsströme.} & de\\
    \hline
    \end{tabular}
    }
    \caption{Some examples of translation by trained SD in the Triangle, in which ablating LS layers from CLL. The long sentence of supervised translation has degeneration compared with short sentences but is kept in the correct direction. These zero-shot translations are biased to supervised directions.}
    \label{tab9}
\end{table*}

\end{document}